\RequirePackage[hyphens]{url}

\documentclass[conference,10pt]{IEEEtran}

\usepackage{cite}

\ifCLASSINFOpdf
  \usepackage[pdftex]{graphicx}     \graphicspath{{./figs/}}
      \DeclareGraphicsExtensions{.pdf,.jpeg,.png}
\else
                  \fi

\usepackage[cmex10,fleqn]{amsmath}
\usepackage[fleqn]{mathtools}
\usepackage{amssymb}
\usepackage{algorithm}
\usepackage{algpseudocode}

\usepackage{array}

\usepackage[labeled,resetlabels]{multibib}
\newcites{S,P}{Supplementary References,Postscript References}

\usepackage{float} 
\usepackage{ulem}
\usepackage{color}
\usepackage{soul}

\usepackage{fancyhdr}

\usepackage{colortbl}

\usepackage{afterpage} 
\usepackage{fixltx2e} 
\usepackage{randtext}
\catcode`\_=11\relax
\newcommand\email[1]{\_email #1\q_nil}
\def\_email#1@#2\q_nil{  \href{mailto:#1@#2}{{\randomize{#1}\emailampersat \randomize{#2}}}}

\newcommand\emailampersat{{\small@}} \catcode`\_=8\relax

\usepackage{xfrac}

\usepackage{picins} 
\usepackage{bm}
\usepackage[hyphenbreaks]{breakurl} \PassOptionsToPackage{hyphens}{url}\usepackage{hyperref}
\usepackage{xcolor}
\hypersetup{
    colorlinks,
    linkcolor={red!50!black},
    citecolor={blue!50!black},
    urlcolor={blue!80!black}
}

\usepackage{textcase}

\usepackage[font=small,labelfont=bf]{caption}

\usepackage[para]{footmisc}

\newcommand{\mystrut}{\vrule height 3ex depth 1.1ex width 0pt }
\newcommand{\mystrutT}[1]{\vrule height #1 depth 1.1ex width 0pt }
\newcommand{\mycenter}[2]{\begin{minipage}[t][#2][c]{0.3in}     \centering {#1}
\end{minipage}}

\usepackage{tikz}
\usetikzlibrary{calc,fit,arrows,shapes.misc}
\newcommand\marktopleft[1]{    \tikz[overlay,remember picture] 
        \node (marker-#1-a) at (0,1.5ex) {};}
\newcommand\markbottomright[1]{    \tikz[overlay,remember picture] 
        \node (marker-#1-b) at (0,0) {};    \tikz[overlay,remember picture,thick,red!100,dashed,inner sep=3pt]         \node[draw, rectangle,rounded corners=1pt,fit=(marker-#1-a.center) (marker-#1-b.center)] {};}
\newcommand\markarrowtopleft[1]{    \tikz[overlay,remember picture] 
        \node (marker-#1-a) at (0,0ex) {};}
\newcommand\markarrowbottomright[1]{    \tikz[overlay,remember picture] 
        \node (marker-#1-b) at (0,0) {};	\tikz[overlay,remember picture,thick,red!100,dashed] \draw[-open triangle 45] ($(marker-#1-a.south)+(+0.07,0.06)$) -- ($(marker-#1-b.south)+(0.09,0.06)$);}
\tikzset{
    actor/.style={
        rectangle, minimum size=6mm, very thick,
        draw=black, top color=white, bottom         color=white!50!black!20     },
    actor crossed out/.style={
        actor, 
        append after command={
            [every edge/.append style={
                thick,
                red!50!black!50,
                shorten >=\pgflinewidth,
                shorten <=\pgflinewidth,
            }]
           (\tikzlastnode.north west) edge (\tikzlastnode.south east)
           (\tikzlastnode.north east) edge (\tikzlastnode.south west)
        }
    }
}

\usepackage{spreadtab}
\STautoround*{2} 
\usepackage{numprint}
\npthousandsep{,}\npthousandthpartsep{}\npdecimalsign{.}

\def\por1{\partial}

\DeclareMathOperator*{\argmax}{\,argmax}

\newcolumntype{M}{>{\centering\arraybackslash}m{\dimexpr0.50\linewidth-2\tabcolsep}}
\newcolumntype{N}{>{\centering\arraybackslash}m{\dimexpr0.10\linewidth-2\tabcolsep}}
\newcolumntype{Y}{>{\raggedleft\arraybackslash}X}

\usepackage{caption}
\DeclareCaptionFormat{myformat}{#1#2#3\hrulefill}
\def\hreftwo#1{\href{#1}{#1}}

\newcommand\litem[1]{\item{\bfseries #1:}}

\begin{document}
\bstctlcite{IEEEexample:BSTcontrol} 
\title{\bf\Large A Multiple-Expert Binarization Framework for Multispectral Images} 
\author{\IEEEauthorblockN{\small Reza \MakeTextUppercase{Farrahi Moghaddam}
\and Mohamed \MakeTextUppercase{Cheriet}}
\IEEEauthorblockA{Synchromedia Lab and CIRROD, ETS (University of Quebec)}
\IEEEauthorblockA{Montreal, QC, Canada H3C 1K3}
\IEEEauthorblockA{Email: \email{imriss@ieee.org}, LinkedIn: \url{https://www.linkedin.com/in/rezafm}} }

\maketitle

\begin{abstract}
In this work, a multiple-expert binarization framework for multispectral images is proposed. The framework is based on a constrained subspace selection limited to the spectral bands combined with state-of-the-art gray-level binarization methods. The framework uses a binarization wrapper to enhance the performance of the gray-level binarization. Nonlinear preprocessing of the individual spectral bands is used to enhance the textual information. An evolutionary optimizer is considered to obtain the optimal and some suboptimal 3-band subspaces from which an ensemble of experts is then formed. The framework is applied to a ground truth multispectral dataset with promising results. In addition, a generalization to the cross-validation approach is developed that not only evaluates generalizability of the framework, it also provides a practical instance of the selected experts that could be then applied to unseen inputs despite the small size of the given ground truth dataset. 
\end{abstract}


\IEEEpeerreviewmaketitle

\section{Introduction}
\label{sec_introduction}
Digitization and computer-based archiving of ancient manuscripts has been of great interest to bring the documented human heritage to the public access and also to produce intangible replications of this heritage in order to preserve it beyond the limits of its physical carriers \cite{Klijn2008,Hebert2006}, \citeS{Lettner2009,Gargano2014,Morales2014}.\footnote{Because of the limited space, the images and some citations are provided in the Supplementary Material and in the Postscript, which is accessible at\\ \hreftwo{http://arxiv.org/pdf/1502.01199.pdf\#page=6}.} Multispectral (MS) imaging has been used toward these goals considering its high capability to record data beyond what is `visible' to human eye \cite{Hedjam2013,Lettner2008,Tonazzini2010,Hollaus2014}. An MS image could be imagined as a generalized color image with more than three bands. However, in practice this may not accurately hold, and there is a big difference between a color image generated by broadband (with FWHM\footnote{FWHM: Full width at half maximum \cite{Wang2014a}.} $> 60\ \text{nm}$) filters calibrated to reproduce the same visual sensation for human eye compared to an MS image generated by a series of intermediate-band ($\text{FWHM} \sim 11-60\ \text{nm}$) or narrowband ($\text{FWHM} \sim 4-10\ \text{nm}$) filters. In addition, various challenges are associated with the MS imaging such as i) nonlinear misregistration among bands, ii) high IR noise, and iii) bigger amount of data.

\begin{figure}[!thb]
\centering
\centerline{\begin{tabular}{@{}c@{}}
\includegraphics[width=3.6in]{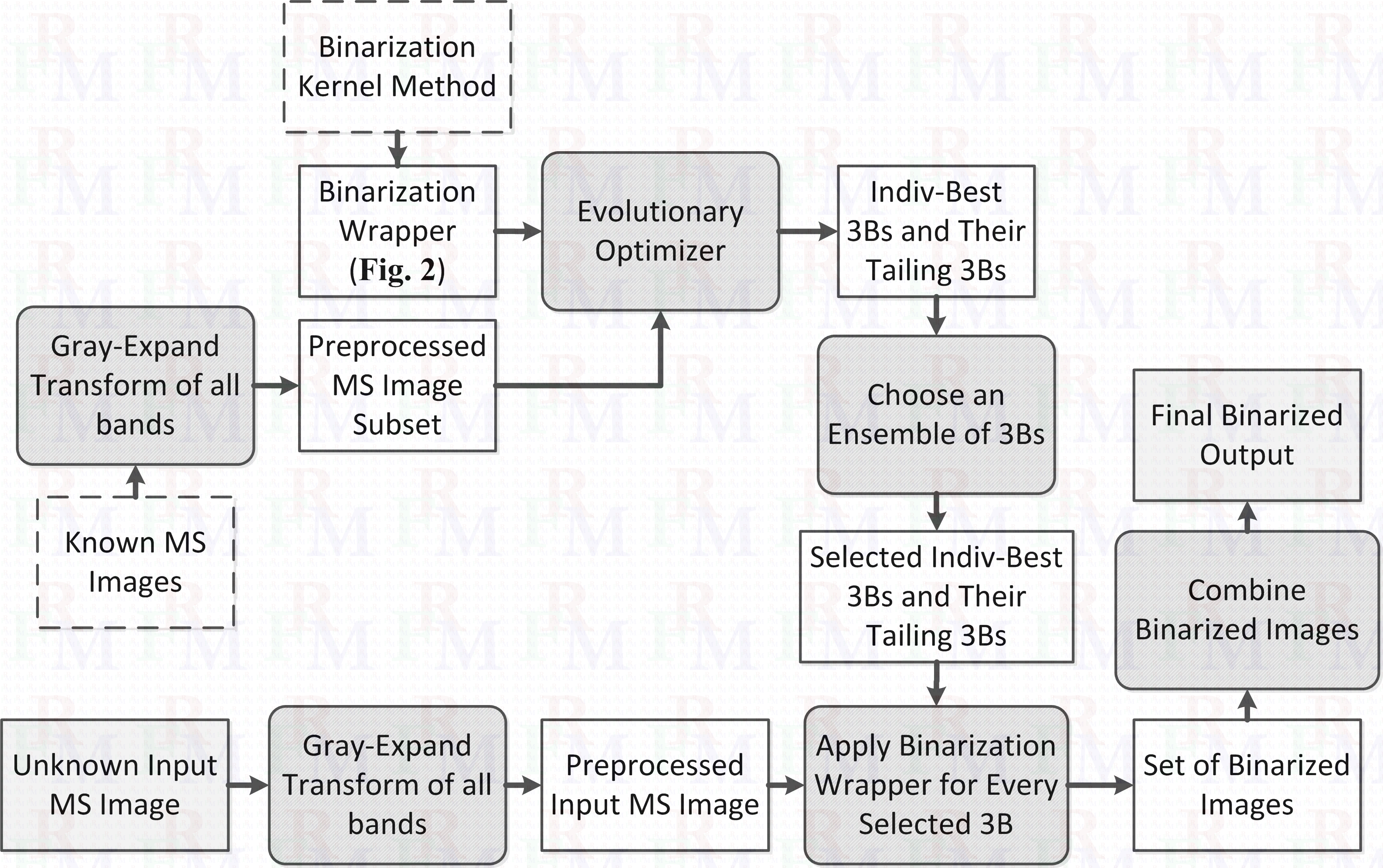}
\end{tabular}}
\caption{The proposed multiple-expert binarization framework.}
\label{fig_Bin_framework_MS1}
\end{figure}

\begin{figure}[!thb]
\centering
\centerline{\begin{tabular}{@{}c@{}}
\includegraphics[width=3.4in]{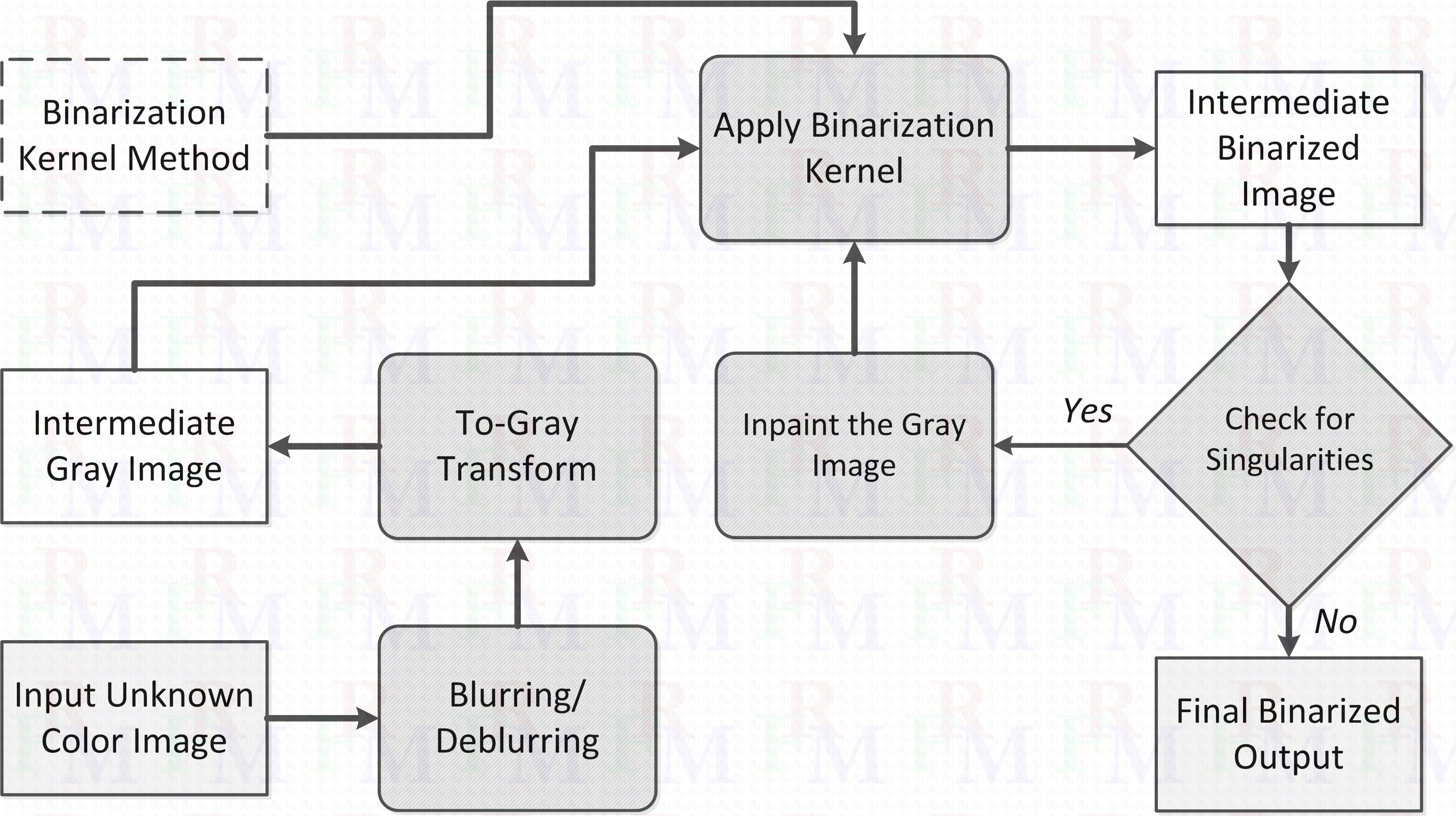}
\end{tabular}}
\caption{A binarization wrapper for any given binarization kernel.}
\label{fig_Bin_wrapper_MS1}
\end{figure}

Segmentation and binarization of MS images could stand as a convergent point between MS image processing and well-studied color/gray document image processing. A great obstacle in this direction is the labor cost associated with creating reference data, especially considering high volume of data contained in MS images. This has been resulted in indirect evaluations of the performance of enhancement and segmentation methods of MS image using OCR or other goal-oriented approaches \cite{Hollaus2014}. With the availability of ground-truth datasets of MS images of ancient manuscripts \cite{Hedjam2013}, developing direct binarization methods of MS images has been pursued in a more systematic way. In this work, a multiple-expert framework to binarize MS images is proposed. The framework uses a subspace selection of MS bands along with a given state-of-the-art  gray-level binarization method, which we call the {\em kernel}. In order to limit the scope of the framework, it is assumed that the kernels are smart enough to adjust their internal parameters for each individual input gray-level image that they receive. In the future, the framework is extended to include optimal selection of the internal parameters the kernels along with the band selection. The framework is multiple expert in the sense that it considers various instances of subspaces in the form of an ensemble of experts, and then simply combines their results. More complex ways of combining the opinions of the multiple experts, such as Unsupervised Ensemble of Experts Reduction (UEoER) approach \cite{Farrahi2013b} and its possible quality-aware generalizations, are not considered in this work.

Each expert will be associated with a set of selected bands and also the way these bands are converted to a single-band (a gray-level) image. In this work, we assume that every subspace has a dimension of 3, i.e., three bands are selected by every expert, and it also assumed that the selected 3-band images are converted in a gray-level image using the traditional gray conversion of the RGB color space \cite{CIE20072007}.\footnote{In the future, other color-to-gray methods, such as the Dual Transform \cite{Farrahi2013d}, will be also considered.
Some implementations can be found here: \href{http://www.mathworks.com/matlabcentral/fileexchange/27578-universal-color-to-gray-conversion}{http://www.mathworks.com/matlabcentral/fileexchange/27578-universal-color\\-to-gray-conversion}.
}

The paper is organized as follows. 
In Section \ref{sec_Basic_Definitions}, the notations and also two kernel binarization methods used in the experiments are provided. The main framework and its components are described in Section \ref{sec_Multiple_3Band_Expert_Binarization_Framework}. The generalized cross-validation approach is presented in Section \ref{sec_CrossValidation_Search_CVS} followed by the experimental results of Section \ref{sec_Experimental_Results_Discussions}. Finally, the conclusions and some prospects for the future are listed in Section \ref{sec_conclusion}.

\section{Basic Definitions}
\label{sec_Basic_Definitions}
In this section, some of the basic concepts are first defined: 
\begin{LaTeXenumerate} \litem{Multispectral Image} A multispectral (MS) image in this work is composed of 8 bands recorded using a multispectral camera (DTA s.r.l. Chroma C3). The bands are produced using a series of filters at 340 nm (florescence), 450 nm (blue), 550 nm (green), 650 nm (red), 800 nm, 900 nm, 1,000 nm, and 1,100 nm. The visible filters are broadband at $\text{FWHM} = 80\ \text{nm}$, while the infrared (IR) filters are intermediate-band at $\text{FWHM} = 50\ \text{nm}$. The camera sensor is a two-phase full-frame low-dark-current CCD (KAF-6303E). Each band of an MS image is recorded in the BW01 protocol, i.e., the black has a value of 0, and white corresponds to 1 \cite{Farrahi2010}. An MS image is denoted by a $N_\text{band}$-tuple $u$, $u = \left(u_i(x)\right)_{i=1}^{N_\text{band}}$, where $i=1,\cdots,N_\text{band}$ is the band counter, $N_\text{band}(=8)$ is the number of spectral bands, and $x$ is a pixel on the image domain $\Omega \subset \mathbb{R}^2$, $u: \Omega \rightarrow \left[0, 1\right]^{N_\text{band}}$.
\litem{Gray-level Image} In this work, a gray-level image is represented by $I$ in general, and it is assumed that it follows the BW10 protocol \cite{Farrahi2010} (black is 1 and white is 0): $I:\Omega \rightarrow \left[0, 1\right]$.
\end{LaTeXenumerate}

\subsection{Binarization Methods}
\label{sec_Binarization_Methods}
Two state-of-the-art binarization methods are considered in this work as the kernel binarization methods:
\begin{LaTeXenumerate} \small
\litem{Laplacian Energy \cite{Howe2012} (LE in short)} The Laplacian-energy method, inspired by a Markov random field model, defines the binarization as a minimization problem for a global energy function. The fidelity term is defined based on the intensity Laplacian that is highly contrast- and intensity-independent. Moreover, the edge information is used to ensure that the binarization boundaries are aligned with edges. Optimization of four internal parameters is considered in the Laplacian-energy method: The two hysteresis thresholds of the Canny edge map, the radius of the Gaussian filter, and the mismatch penalty.
\litem{Phase Congruency \cite{Ziaei2014} (PC in short)} This method uses a combination of phase feature maps, such as the maximum moment of phase congruency covariance (MMPCC) and the local weighted mean phase angle (LWMPA) and the regional minima feature maps, and also adaptive Gaussian and median filtering in order to provide a robust and consistent binarization performance for various types of degradation. The method has three explicit internal parameters: The number of scales, the number of orientations of the wavelet transform, and also the threshold of noise  standard deviation. In this work, optimization of these parameters for every input image is not considered, and a fixed set of optimal values for these parameters obtained using DIBCO series datasets \cite{Pratikakis2013} is used. The adaptability of the internal processes is main capability of this method to optimize its performance across various degradation types.
\end{LaTeXenumerate}

\section{The Dataset}
\label{sec_Dataset}
One of the ground truth datasets of MS images provided in \cite{Hedjam2013} is used in this work.\footnote{Available online: \url{http://www.synchromedia.ca/databases/msi-histodoc}} The dataset composed of 21 MS images of ancient manuscripts, which follow the description of MS images provided in Section \ref{sec_Basic_Definitions}. Every MS image is accompanied with a binary image that provides segmentation of the text on the associated manuscript image. 

\section{The Proposed Multiple-Expert Framework}
\label{sec_Multiple_3Band_Expert_Binarization_Framework}
The schematic diagram of the framework is shown in Figure \ref{fig_Bin_framework_MS1}. The framework receives a set of {\em given} ground truth MS images and a gray-level binarization method (the kernel). The spectral bands of the MS images are first enhanced using the Gray-Expand transform \cite{Farrahi2012} in order to increase the differentiability of the textual information. The kernel binarization is then wrapped (see Section \ref{sec_Binarization_Wrapper_Method}), and using an evolutionary optimizer \cite{Farrahi2012g}, the individual-best 3Bs of every MS image along with their {\em tailing} suboptimal 3Bs are obtained (see Section \ref{sec_Optimization_Individual_Best_3Bs}).  Then, the rare-or-frequent 3Bs are chosen to create an ensemble of 3B experts (See Section \ref{sec_Selection_Experts}). Having such an ensemble, any new unseen MS image is passed through the preprocessing and then through the binarization wrapper for every member of the ensemble, and the final binarization output is obtained by combining all the binary outputs in a simple averaging step. The details of various steps of the framework are presented in the following subsections.

\subsection{The Proposed Binarization Wrapper Method}
\label{sec_Binarization_Wrapper_Method}
The wrapper adds three features to every kernel that it hosts: i) It passes individual bands of the input color/multiband image through a blurring/deblurring process in order to minimize the registration/mismatching error among bands, ii) , and iii) after obtaining the output of the kernel on the gray-level image, it performs a test to ensure the kernel is not trapped on some small regions of the input image.

For blurring and deblurring steps, a Gaussian profile with $\sigma=0.5$ and a radius of 5 pixels and another Gaussian profile with $\sigma=5$ and a radius of 5 pixels are respectfully used. The luminance color-to-gray transform is considered in this work, and the singularity test is performed using a ratio threshold followed by an inpainting step if necessary.

\subsection{Optimization and Individual-Best 3Bs}
\label{sec_Optimization_Individual_Best_3Bs}
In order to obtain the best 3-Bands (3B) selection associated with each one of MS image in the given dataset, an evolutionary optimizer, called Curved Space Optimizer (CSO)  \cite{Farrahi2012g}, is considered. In the process to obtain the global best 3B, the optimizer visits various 3B values. In this work, in addition to the global optimal (the best) 3B associated with an MS image (and implicitly with a specific kernel method), the top tailing 3Bs of that image are also reported. An example of the output of the optimizer for the image `z30' of the dataset is provided in Table \ref{tab_optimizer_output_z30_1} (considering the LE method as kernel). 

\begin{table}[!htb]
\small
\centering
\begin{tabular}{||l||c|c|c||}\hline\hline
$\text{Optimality}{\mkern -6mu}\setminus{\mkern -6mu}\text{Bands}$ & $\text{Band}_\text{R}$ & $\text{Band}_\text{G}$ & $\text{Band}_\text{B}$ \\\hline\hline
Global (Individual-Best) & 8 & 2 & 1 \\\hline\hline
Sub-Optimal$_1$ & 6 & 2 & 6 \\\hline
Sub-Optimal$_2$ & 7 & 2 & 6 \\\hline
Sub-Optimal$_3$ & 5 & 3 & 2 \\\hline
\hline
\end{tabular}
\caption{The global optimal and also sub-optimal (tailing) 3Bs of the image `z30' in the given dataset.}
\label{tab_optimizer_output_z30_1}
\end{table}

\subsection{Selection of the Experts}
\label{sec_Selection_Experts}
Having all individual-best and {\em tailing} 3Bs of every MS image of a given (sub-)dataset, the following procedure is considered to extract the rare and also the frequent 3Bs in order to build the set of multiple experts required by the framework. First, for every image, the ranked list of the 3Bs are descendingly weighted with integer numbers starting from 0. Then, all the weighted 3Bs of all images are aggregated using the summation function. The rare instances are collected by choosing those 3Bs that have a sum of zero. In addition, a number of most frequent 3Bs are selected by choosing those that have the lowest negative sum (i.e., the highest absolute sum). In this work, we consider up to a number of 5 most frequent 3Bs. The odd-sized union of rare and frequent 3Bs sets is considered to get the final set of selected experts.

\section{Cross-Validation Search (CVS) Extension}
\label{sec_CrossValidation_Search_CVS}
In the experimental section, Section \ref{sec_Experimental_Results_Discussions}, we will use the cross validation approach to evaluate how much the proposed framework is generalizable considering the small size of the ground truth data. However, the next challenge would be how to select a proper subset of data that can be used to process unseen, not-yet-available data (probably the data of an upcoming contest), especially when the ground truth data is small that is the case in this work.\footnote{When the size of available data is small, cross-validation approaches are common in order to validate a methodology \cite{Konen2010} or a hypothesis \cite{Knaub1982}.} Here, we first discuss this challenge, and then we propose another approach to perform such a selection in a fair way, i.e., maximizing the performance while avoiding possible over fitting. The proposed extension is called the Cross-Validation Search (CVS). 

Let us review the notation of a $p$-holdout cross validation ($0 < p < 1$). Assuming that the `given' ground truth data is of a total size of $N$, the `training' data used in every iteration of the cross-validation process would be a randomly selected subset of the given data with a size of $(1-p)N$. The rest of the data, i.e., a subset with a size of $pN$ would play the role of the `validation' data in that particular iteration. If a number of $N_\text{CV}$ iterations is performed, the mean and the standard variation values of a measure, such as the mean F-measure on the validation data, could be used for the purpose of validating whether the method under study is generalizable. We will follow this procedure to validate the proposed framework.

The next question would be how to choose a subset of `given' data to be used in processing unseen, upcoming new data. Various strategies could be used: 
i) Minimal standard variation on the validation subset: Although this well-known approach is fair and {\em implicitly} searches for the most generalizable  set by selecting the {\em easiest} validation subset, it could default on itself when the size of the given data is small,
ii) Maximal performance on the validation data: There is again a high chance of low performance especially because usually a $p<0.5$ is selected,
iii) Maximal performance on the whole given data: Here, there is a high risk of over fitting, and,
iv) Using the whole given data as the training data ($p=0$): There is a chance of both over fitting and also low performance even on the given data, especially in the case of multi-expert methods. The last approach is denoted {\em All 3Bs} in this work.

Here, we propose to use an extension, called Cross-Validation Search (CVS), in the form of a cross-validation measure limited to the validation subset in order to avoid over fitting while searching for maximum performance. To define such a measure, we assume that, for each member of the given data, the F-measure scores of {\em three} experts are available: i) a shared `typical' expert, ii) an upper-bound expert, and iii) the multiple-expert method under study. It is worth mentioning that there is probably a different upper-bound (individual-best) expert for each member of the given set. In this work, which is limited to 3-band subspace experts, the shared typical expert is assumed to be the trivial RGB-band expert, and the individual-best experts are also simple 3-band experts (without any combination of 3Bs). The three average performance of the three experts is then is calculated on the validation subset: 
\begin{equation}
\widehat{\text{FM}}_{k} = \left(\widehat{\text{FM}}_{\text{typ},k}, \widehat{\text{FM}}_{\text{bes},k}, \widehat{\text{FM}}_{\text{mul},k}\right),
\end{equation}
where $\widehat{\text{FM}}_{\text{typ},k}$, $\widehat{\text{FM}}_{\text{bes},k}$, and $\widehat{\text{FM}}_{\text{mul},k}$ are the average F-measure performance of the typical expert, the individual-best expert(s), and the multiple-expert under study on the validation subset of a particular iteration $k$, respectively.
The proposed CVS measure of the iteration $k$ is then defined as follows:
\begin{align}  \widehat{\text{CVS}}_{k}   := &  \left(\widehat{\text{FM}}_{\text{mul},k} - \widehat{\text{FM}}_{\text{typ},k}\right) -  \left(\widehat{\text{FM}}_{\text{bes},k} - \widehat{\text{FM}}_{\text{mul},k}\right), \label{e_CVS_1_1}\\
   = & 2 \widehat{\text{FM}}_{\text{mul},k} - \widehat{\text{FM}}_{\text{typ},k} -  \widehat{\text{FM}}_{\text{bes},k}.
\end{align}
The first term in Equation (\ref{e_CVS_1_1}) represents how much the method is better than {\em typical} expert, while the second term measures its {\em perfectness}. Therefore, the whole $\widehat{\text{CVS}}_{k}$ calculates the goodness of a training subset on its associated validation subset {\em relative} to the upper and lower bounds given by the typical and individual-best experts. Instead of a ratio, the {\em difference} is used in order to avoid sensitivity to small improvements. It could be argued that a training subset $k$ which provides a high $\widehat{\text{CVS}}_{k}$ value would also have a good performance on itself.
In the proposed CVS approach, the particular training subset associated with the iteration with highest $\widehat{\text{CVS}}_{k}$ is selected to build the final multiple-expert method for upcoming inputs:
\begin{equation}
k^* = \argmax_{k=1,\cdots, N_\text{CV}} \widehat{\text{CVS}}_{k} 
\end{equation}
The 3Bs associated to the members of the training subset of iteration $k^*$ are used to build a multiple-expert method with an estimated CVS performance of $\widehat{\text{CVS}}_{k^*}$.

It is worth mentioning that we carry out the iteration process twice. First time, it is used to validate a method under study the same way it is performed in a standard cross-validation process, and in each iteration a pure random selection is used. The second time, it calculates the optimal iteration $k^*$, and instead of using a pure random selection, we use the same heuristics optimizer algorithm of \cite{Farrahi2012g} to control the selection process of training and validation subsets. For the purpose of simplicity of the notation, the same number of iteration is used for this rerun. We argue that the high number of possible selections and also the low number of the iterations performed would lead a pure random selection process to settle with a sub-optimal or even non-optimal result. Using an optimizer could be imagined as setting $N_\text{CV} \rightarrow \infty$. However, it should be again mentioned that the average statistics of the cross validation, provided in the first row of Table \ref{tab_CV_20P_scores1} in the next section, are calculated using a completely random selection with $N_\text{CV}=50$, and no optimization was performed.

\section{Experimental Results and Discussions}
\label{sec_Experimental_Results_Discussions}
In Table \ref{tab_indbest_all3bs_RGB_scores1}, the performance of the binarization wrapper introduced in Section \ref{sec_Binarization_Wrapper_Method} is presented using the LE binarization method as the kernel. First, using the evolutionary optimization algorithm of \cite{Farrahi2012g}, the best 3B is determined for every multispectral image in the dataset. The performance of these individual-best 3Bs on the whole dataset is provided in the first row of the table. The $\text{FM}_\text{avg}$, $\text{FM}_\text{std}$, $\text{FM}_{\text{avg}, 1}$, and $\text{FM}_{\text{std}, 1}$ are the average of the F-measure, the standard deviation of the F-measure, the average of the F-measure excluding the worst image, and the standard deviation of the F-measure excluding the worst image, respectively. For the purpose of comparison and also to have a `typical' way of selecting the 3 bands, the case of RGB bands is provided in the second row. As can be seen, there is a big difference (around 10\%) between their performance. In the third row, labeled All 3Bs, the performance of the proposed multiple 3Bs framework is provided in a case where the best 3B of all given images along with their tailing best 3Bs are combined as described in Section \ref{sec_Selection_Experts}. 
As discussed in Section \ref{sec_CrossValidation_Search_CVS}, the performance of the All 3Bs case could not be guaranteed to be generalizable.

\begin{table}
\small
\centering
\begin{tabular}{||l||c|c|c|c||}\hline\hline
$\text{Case}{\mkern -5mu}\setminus{\mkern -5mu}\text{Measure}$ & $\text{FM}_\text{avg}$ & $\text{FM}_\text{std}$ & $\text{FM}_{\text{avg}, 1}$ & $\text{FM}_{\text{std}, 1}$ \\\hline\hline
Individual Best &  \marktopleft{c3}80.81 & 5.37 & 81.49 & 4.48 \\\hline
RGB Bands & 69.58\markbottomright{c3} & 19.91 & 72.30 & 15.90 \\\hline
All 3Bs & 73.23 & 14.66 & 75.86 & 8.54 \\\hline
\hline
\end{tabular}
\caption{The performance of the individual-best 3B binarization wrapper with the LE method as the kernel, and also that of all the 3Bs of all 21 MS images combined using the method of Section \ref{sec_Selection_Experts}.}
\label{tab_indbest_all3bs_RGB_scores1}
\end{table}

\begin{table}
\small
\centering
\begin{tabular}{||l||c|c|c|c|c||}\hline\hline
$\text{Case}{\mkern -5mu}\setminus{\mkern -5mu}\text{Measure}$ & $\text{FM}_\text{avg}$ & $\text{FM}_\text{std}$ & $\text{FM}_{\text{avg}, 1}$ & $\text{FM}_{\text{std}, 1}$ & \multicolumn{1}{>{\mystrut}c||}{$\widehat{\text{CVS}}$} \\\hline\hline
Average CV$^\dagger$ & 77.19 & 6.50 & 80.14 & 2.73 & -0.05 \\\hline\hline
$k^*$ $^\dagger$ & \marktopleft{c1}72.23\markarrowtopleft{a1} & 17.07 & 80.40 & 6.02 & \markarrowbottomright{a1}\marktopleft{c2}3.65\markbottomright{c2} \\\hline
RGB Bands$^{\dagger,\ddagger}$ & {\em 62.38} & 32.25 & - & - & - \\\hline
Individual Best$^{\dagger,\ddagger}$ & 78.44\markbottomright{c1} & 8.47 & - & - & - \\\hline\hline
$\mathbf{k^*}$ & {\bf 73.93} & {\bf 14.05} & {\bf 76.46} & {\bf 8.91} & {\bf 3.65} \\\hline\hline
All 3Bs & 73.23 & 14.66 & 75.86 & 8.54 & - \\\hline
RGB Bands & {\em 69.58} & 19.91 & 72.30 & 15.90 & - \\\hline
Individual Best & 80.81 & 5.37 & 81.49 & 4.48 & - \\\hline
\hline
\end{tabular}
\caption{The performance of the CVS approach ($p=0.2$). Notes: $^\dagger$The performance presented is associated with the `validation' subset. 
$^\ddagger$This performance is associated with the $k^*$ iteration.}
\label{tab_CV_20P_scores1}
\end{table}

\begin{table}
\small
\centering
\setlength{\tabcolsep}{5pt}
\begin{tabular}{||l||c|c|c|c||c|c||}\hline\hline
 & & \multicolumn{3}{c||}{Proposed CVS} & \multicolumn{2}{c||}{Minimum Std} \\\hline
$p{\mkern -5mu}\setminus{\mkern -5mu}\text{Measure}$ & \#$^\dagger$ & $\text{FM}_\text{avg}$ & $\text{FM}_\text{std}$ & \multicolumn{1}{>{\mystrut}c||}{$\widehat{\text{CVS}}_{k^*}$} & $\text{FM}_\text{avg}$ & $\text{FM}_\text{std}$ \\\hline\hline
$p=0.10$ & 19 & 73.27 & 15.05 & 10.02 & 73.48 & 14.65 \\\hline
$p=0.20$ & 17 & {\bf 73.93} & 14.50 & 3.65 & 73.57 & 14.53 \\\hline
$p=0.50$ & 11 & {\bf 74.82} & 11.61 & 3.59 & 73.49 & 14.35 \\\hline
$p=0.90$ & 3 & {\bf 75.05} & 13.40 & 2.41 & 74.49 & 13.24 \\\hline
$\mathbf{p=0.97}$ & 1 & {\bf 76.32} & 11.05 & 0.98 & 74.01 & 11.64 \\\hline
\hline
\end{tabular}
\caption{The performance of the proposed CVS approach against that of the minimal standard deviation across the parameter $p$  (the LE method is the kernel). Note: $^\dagger$This column denotes the number of sample images used in the training subset.}
\label{tab_CV_MultipleP_scores1}
\end{table}

The results obtained using the proposed CVS approach are provided in Tables \ref{tab_CV_20P_scores1} and \ref{tab_CV_MultipleP_scores1} in comparison with those of minimum standard deviation approach. In Table \ref{tab_CV_20P_scores1}, for the case of $p=0.2$, the $k^*$ iteration achieved a performance comparable to that of the All 3Bs case. Interestingly, in Table \ref{tab_CV_MultipleP_scores1}, an improved performance of $\Delta \text{FM}_\text{avg}=1.33\%$ compared to the minimum standard deviation approach was achieved with a smaller training subset size ($p=0.5$). It is worth noting that the case of $p=0.97$ has only 1 image in the training subset (resulted in a multiple-expert of five 3Bs).

\begin{table}[!htbp]
\small
\centering
\setlength{\tabcolsep}{3pt} \begin{tabular}{||l|c|c|c|c||l|c|c|c|c||}\hline\hline
 & & \multicolumn{3}{c||}{CVS} & & & \multicolumn{3}{c||}{CVS} \\\hline
$p$ & \#$^\dagger$ & $\text{FM}_\text{avg}$ & $\text{FM}_\text{std}$ & \multicolumn{1}{>{\mystrut}c||}{$\widehat{\text{CVS}}_{k^*}$} & $p$ & \# & $\text{FM}_\text{avg}$ & $\text{FM}_\text{std}$ & \multicolumn{1}{>{\mystrut}c||}{$\widehat{\text{CVS}}_{k^*}$} \\\hline\hline
$0.10$ & 19 & 76.72 & 7.85 & 3.43 & $0.60$ & 9 & 76.38 & 8.34 & -0.24 \\\hline
$0.20$ & 17 & 76.70 & 7.82 & 1.35 & $0.90$ &   3 & 76.99 & 7.22 & 0.60 \\\hline
$0.50$ & 11 & 76.54 & 7.58 & 0.36 & $\mathbf{0.97}$ &   1 & {\bf 77.29} & 7.20 & -0.56 \\\hline\hline
 & & \multicolumn{3}{c||}{Individual Best} & & & \multicolumn{3}{c||}{RGB Bands} \\\hline
$p$ & \# & $\text{FM}_\text{avg}$ & $\text{FM}_\text{std}$ & \multicolumn{1}{>{\mystrut}c||}{$\widehat{\text{CVS}}_{k^*}$} & $p$ & \# & $\text{FM}_\text{avg}$ & $\text{FM}_\text{std}$ & \multicolumn{1}{>{\mystrut}c||}{$\widehat{\text{CVS}}_{k^*}$} \\\hline
- & 21 & 79.69 & 6.36 & - & - & 1 & {\em 74.80} & 9.86 & - \\\hline
\hline
\end{tabular}
\caption{The performance with the PC method as the kernel. 
}
\label{tab_CV_MultipleP_scores_PC1}
\end{table}

\setcounter{footnote}{\thefootnote+1}

The same procedure was carried out using the PC method as the kernel; the results are reported in Table \ref{tab_CV_MultipleP_scores_PC1}. Less variation across the $p$ values that can be attributed to the more adaptability of the kernel's internal processes. 
The multiple-3Bs binarization methods developed using the proposed framework with $p=0.97$ (considering the LE method or PC method as the kernel method) will be used as baseline methods in the ICDAR 2015 MultiSpectral Text Extraction Contest \cite{Hedjam2015}.\footnotemark[\thefootnote]

\footnotetext[\thefootnote]{ 
\href{http://www.synchromedia.ca/system/files/MSTEx_ICDAR15_CFP.pdf}{http://www.synchromedia.ca/system/files/MSTEx\_ICDAR15\_CFP.pdf},
\href{http://www.synchromedia.ca/competition/ICDAR/mstexicdar2015.html}{http://www.synchromedia.ca/competition/ICDAR/mstexicdar2015.html}.}

\begin{figure}[!thb]
\centering
\setlength{\tabcolsep}{1pt}
\centerline{\begin{tabular}{cl}
\mycenter{(a)}{-12.1ex} & \mystrutT{12.1ex}\includegraphics[width=2.9in]{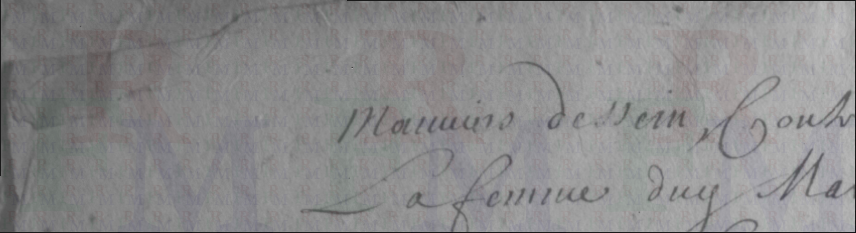} \\
\mycenter{(b)}{-12.1ex}   & \mystrutT{12.1ex}\includegraphics[width=2.9in]{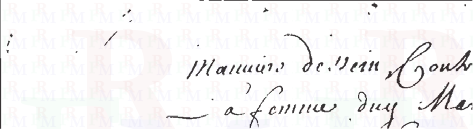} \\
\mycenter{(c)}{-12.1ex}   & \mystrutT{12.1ex}\includegraphics[width=2.9in]{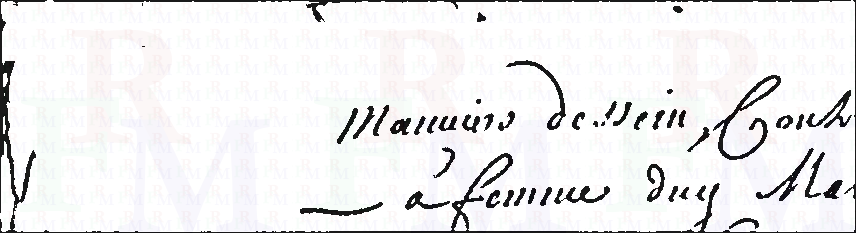} \\
\end{tabular}}
\caption{A subjective evaluation. 
a) z67, $2^{nd}$ dataset \cite{Hedjam2013}.
b) Result of \cite{Mitianoudis2014}.
c) Result of the CVS (p = 0.97, PC).}
\label{fig_Bin_framework1}
\end{figure}

\iftrue
On another dataset of 9 multispectral images from \cite{Hedjam2013}, a $\text{FM}_{\text{avg}, 1}$ score of 78.76 was obtained with the CVS ($p=0.97$, LE) compared to a $\text{FM}_{\text{avg}, 1}$ of 73.40 obtained using RGB-bands and the LE kernel, and to a $\text{FM}_{\text{avg}, 1}$ score of 79.53 obtained by the best reported method \cite{Mitianoudis2014} (see Figure \ref{fig_Bin_framework1}). We also obtained a $\text{FM}_{\text{avg}, 1}$ of 79.01 with the CVS combined with a generalization of the LE method inspired from \cite{Mitianoudis2014}.
\fi

Finally, it is worth mentioning that although the multiple-expert method would explicitly use {\em all} the best set of parameters of every member of the `training' data, it would not {\em actually} perform any optimization or `tuning' toward maximizing the performance on the {\em whole} training data; a multiple-expert way of augmenting the individual-best parameter sets could result in downgraded performance.

\section{Conclusions and Future Prospects}
\label{sec_conclusion}
A binarization framework for multispectral images based on multiple-expert 3-band selection has been proposed. The framework comprised of a binarization wrapper, an optimizer, and an expert selection process. It receives a dataset of ground truth images and a gray-level binarization (kernel), and then generates an ensemble of experts in the form of three-band subspace selections that can be used along the wrapper to binarize any new input image. In addition, a generalized cross-validation approach is introduced to minimize the side-effects of small size of ground truth datasets. The framework and the cross-validation approach have been applied to a ground truth multispectral-image dataset along with two state-of-the-art gray-level binarization methods with promising results.

In future, i) impact of other color-to-gray conversions, ii) more than 3 (and also variable) number of bands, iii) extension of the experts to cover internal parameters of the binarization kernels, and iv) generalized measures (of members instead of averages) for the CVS approach will be considered. Study of the impact of the evolutionary optimization in cross-validation partitioning toward maximizing the CVS measure on other datasets, along with integration of quality-aware ensemble-of-expert reduction approaches to reduce the size of the selected-experts set are other directions to be investigated.

{{\bf Acknowledgment.} The authors thank the NSERC of Canada (under Grant RGPDD/451272-13 and Grant RGPIN/138344-14) and the SSHRC of Canada (under Grant 412-2010-1007) for their financial support.}

\begingroup
\bibliographystyle{IEEEtran} \bibliography{imagep}
\endgroup

\afterpage{\clearpage}
\clearpage

\renewcommand\appendixname{ Supplementary Material }

\appendix

\setcounter{page}{1}
\setcounter{table}{0}
\setcounter{figure}{0}
\setcounter{section}{19}
\renewcommand{\thepage}{S-\arabic{page}}
\renewcommand{\thefigure}{S-\arabic{figure}}
\renewcommand{\thetable}{P-\arabic{table}}
\renewcommand{\thesection}{Supplementary Material \Alph{section}}
\renewcommand{\thesubsection}{Supplementary Material \Alph{section}.\Alph{subsection}}

\section{Supplementary Material}

\subsection{Full Description of the Multispectral Datasets}
\label{sec_suppl_mat_MS_datasets}
Although the dataset used for training and testing the proposed framework has been described in Section \ref{sec_Dataset}, a more detailed description of all datasets involved in this work is provided here.

\subsubsection{The 20MS Dataset \cite{Hedjam2013}}
\label{sec_suppl_mat_MS_datasets_bigDS}
This dataset is the main dataset used in this work, and it is provided in \cite{Hedjam2013}. It contains 20 multispectral (MS) images. Therefore, we call this dataset 20MS in short. The MS images of the 20MS dataset are available to public at the following web page: \url{http://www.synchromedia.ca/databases/msi-histodoc}, under the file name \href{http://www.synchromedia.ca/system/files/S-MS\_1.zip}{S-MS\_1.zip}.\footnote{ \url{http://www.synchromedia.ca/system/files/S-MS\_1.zip}.} The results presented in Tables \ref{tab_CV_20P_scores1}, \ref{tab_CV_MultipleP_scores1}, and \ref{tab_CV_MultipleP_scores_PC1} have been generated using this dataset. 

In the text, the configuration used along the proposed framework is usually denoted in the form of a tuple. For example, when we talk about the CVS ($p=0.97$, LE) configuration, it means that we use the CVS selection, a $p$ value of $0.97$, and the LE kernel along the proposed framework. It is worth mentioning that there is another element in the tuple that has been ignored in the text. This element denotes the dataset used for training. Because we only used the 20MS dataset as the training set, the associated element has been dropped from the tuple. In full form, the example configuration mentioned above would be denoted the CVS ($p=0.97$, LE, 20MS dataset) configuration.

\subsubsection{The 9MS Dataset \cite{Hedjam2013}}
\label{sec_suppl_mat_MS_datasets_smallDS}
Another MS image dataset has been introduced in \cite{Hedjam2013} that contains 9 MS images. We call this dataset 9MS, and in the text it has been referred to in the comparison with the method introduced in \cite{Mitianoudis2014}. The dataset is available at: \url{http://www.synchromedia.ca/databases/HISTODOC1}, under the file name \href{http://www.synchromedia.ca/system/files/HISTODOC1.zip}{HISTODOC1.zip}.\footnote{ \url{http://www.synchromedia.ca/system/files/HISTODOC1.zip}.} 

\subsubsection{The 10MS Dataset \cite{Hedjam2015}}
\label{sec_suppl_mat_MS_datasets_testDS}
Along with the ICDAR 2015 MultiSpectral Text Extraction Contest \cite{Hedjam2015},\footnotemark[4] a separate MS image dataset containing 10 MS images has been developed. Because of the timeline of the contest, the dataset has not been available to public at the time we write this paper. 

\subsubsection{The 3MS Dataset}
\label{sec_suppl_mat_MS_datasets_3imageDS}
In \citeS{Hedjam2014}, another dataset of 3 MS images were introduced for the purpose of invisible text detection, which we call the 3MS dataset. The MS images of the 3MS dataset could be retrieved by contacting the authors of \citeS{Hedjam2014}.

\subsection{Subjective Results and Comparison}
\label{sec_suppl_mat_subj_compar}
Because of the limited space in the main paper, here we provide some examples of the performance of the proposed framework on the images from various datasets. Also, a subjective comparison with the results of previously published work (\cite{Mitianoudis2014} and \citeS{Hedjam2014}) on binarization and invisible text detection in the multispectral document images is presented.

\begin{figure*}[!thb]
\centering
\setlength{\tabcolsep}{3pt}
\centerline{\begin{tabular}{||l|l||}\hline\hline
\mystrutT{17ex}\includegraphics[width=3.5in]{sz67_F2} & 
\includegraphics[width=3.5in]{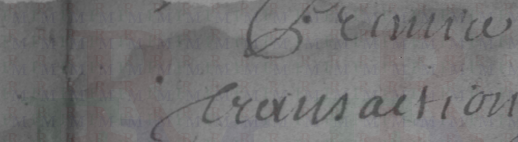} \\
(a) z67, $2^{nd}$ dataset \cite{Hedjam2013} (the second band is shown). & (b) z95, $2^{nd}$ dataset \cite{Hedjam2013} (the second band is shown). \\\hline\hline
\mystrutT{17ex}\includegraphics[width=3.5in]{4_2_d} & 
\includegraphics[width=3.5in]{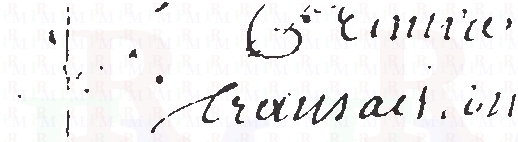} \\
(c) Result of \cite{Mitianoudis2014} on (a); FM=0.65. & (d) Result of \cite{Mitianoudis2014} on (b); FM=0.79. \\\hline\hline
\mystrutT{17ex}\includegraphics[width=3.5in]{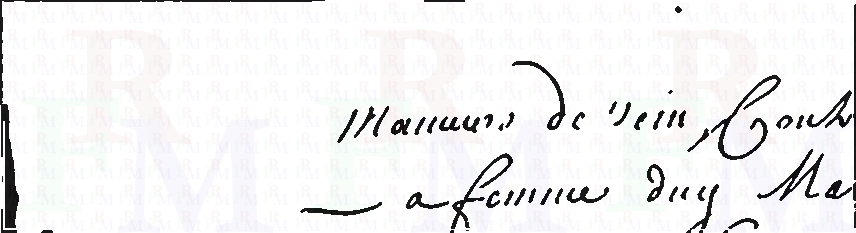} & 
\includegraphics[width=3.5in]{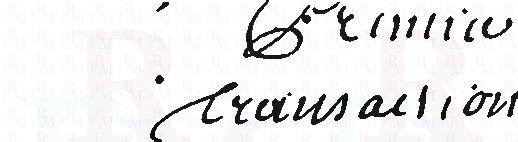} \\
(e) Result of the CVS (p = 0.97, LE) on (a); FM=0.60. & (f) Result of the CVS (p = 0.97, LE) on (b); FM=0.84. \\\hline
\mystrutT{17ex}\includegraphics[width=3.5in]{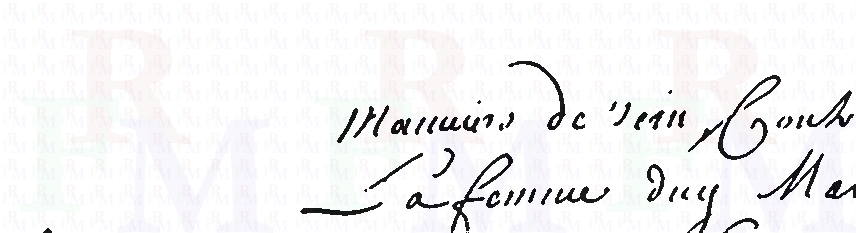} & 
\includegraphics[width=3.5in]{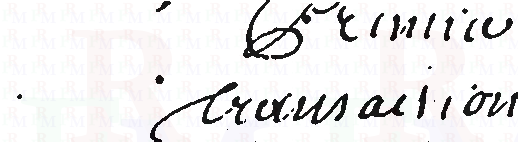} \\
(g) Result of the CVS (p = 0.97, LE + \cite{Mitianoudis2014}) on (a); FM=0.73. & (h) Result of the CVS (p = 0.97, LE + \cite{Mitianoudis2014}) on (b); FM=0.84. \\\hline
\mystrutT{17ex}\includegraphics[width=3.5in]{z67_PC_FullBestOutputSetof5_CVS} & 
\includegraphics[width=3.5in]{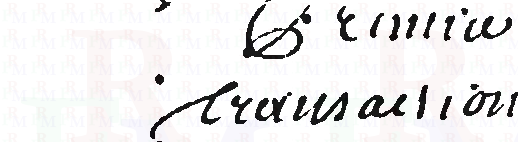} \\
(i) Result of the CVS (p = 0.97, PC) on (a); FM=0.55. & (j) Result of the CVS (p = 0.97, PC) on (b); FM=0.80. \\\hline
\mystrutT{17ex}\includegraphics[width=3.5in]{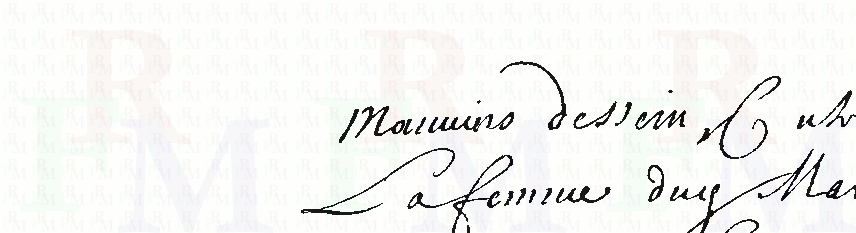} & 
\includegraphics[width=3.5in]{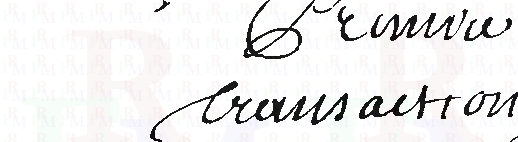} \\
(k) The GT of (a) \cite{Hedjam2013}. & (l) The GT of (b) \cite{Hedjam2013}. \\\hline\hline
\end{tabular}}
\caption{The subjective evaluation of some of the CVS combinations along with the method reported in \cite{Mitianoudis2014} on two images from the second dataset of \cite{Hedjam2013}.}
\label{fig_Bin_framework_Subj1}
\end{figure*}

Figure \ref{fig_Bin_framework_Subj1} provides a visual comparison of various methods on the 9MS dataset of \cite{Hedjam2013} (see Section \ref{sec_suppl_mat_MS_datasets_smallDS} for more information). To be consistent with the results reported in \cite{Mitianoudis2014}, two of the multispectral images, namely z67 and z95, were chosen in this figure. The second band of the input images, the output reported in \cite{Mitianoudis2014}, the output of the proposed framework in CVS ($p=0.97$, LE) configuration, the output of the CVS ($p=0.97$, LE + \cite{Mitianoudis2014}) configuration, the output of the CVS ($p=0.97$, PC) configuration, and the ground truth images are shown in the figure rows, respectively. In particular, it is worth mentioning that some methods receive low scores, especially the CVS ($p=0.97$, PC) configuration, for the input z67 (Figure \ref{fig_Bin_framework_Subj1}(a)) because of a black frame around this image that does not actually exist on the manuscript.

\begin{figure*}[!thb]
\centering
\setlength{\tabcolsep}{3pt}
\centerline{\begin{tabular}{||l|l||}\hline\hline
\mystrutT{11ex}\includegraphics[width=3.5in]{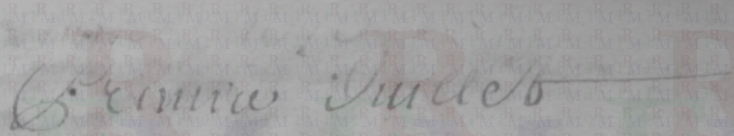} & 
\includegraphics[width=3.5in]{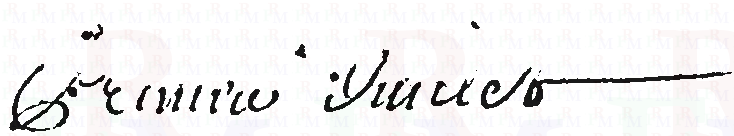} \\
(a) MS image of Fig. 4(a1) in \citeS{Hedjam2014}. & (b) Result of the CVS (p = 0.97, PC) on (a); FM=0.71. \\\hline\hline
\mystrutT{11ex}\includegraphics[width=3.5in]{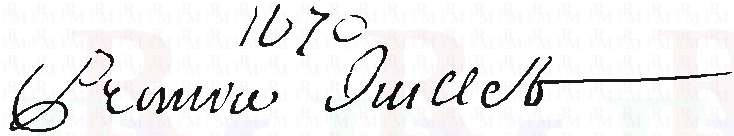} & 
\includegraphics[width=3.5in,decodearray={1 0 1 0 1 0}]{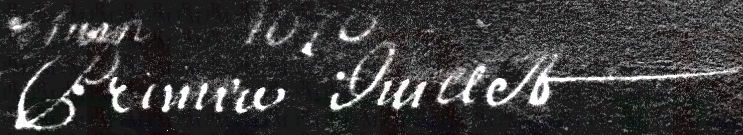} \\
(c) Ground truth of (a); & (d) Result of \citeS{Hedjam2014} on (a); FM=0.44. The image is in BW10. \\\hline\hline
\mystrutT{32ex}\includegraphics[width=3.5in]{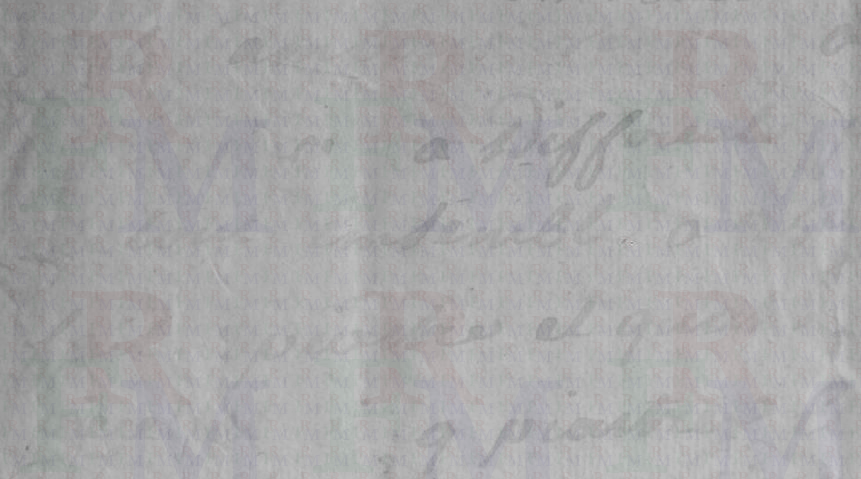} & 
\includegraphics[width=3.5in]{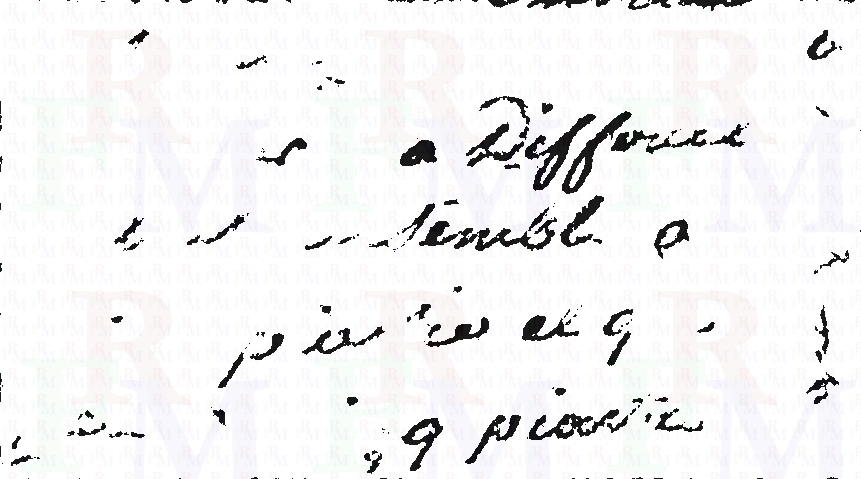} \\
(e) MS image of Figure 4(a2) in \citeS{Hedjam2014}. & (f) Result of the CVS (p = 0.97, PC) on (e); \\\hline\hline
\mystrutT{32ex}
\begin{tikzpicture}
\node [actor crossed out] (b) {\phantom{\includegraphics[width=3.5in]{F4_a2}}};
\end{tikzpicture}  & 
\includegraphics[width=3.5in,decodearray={1 0 1 0 1 0}]{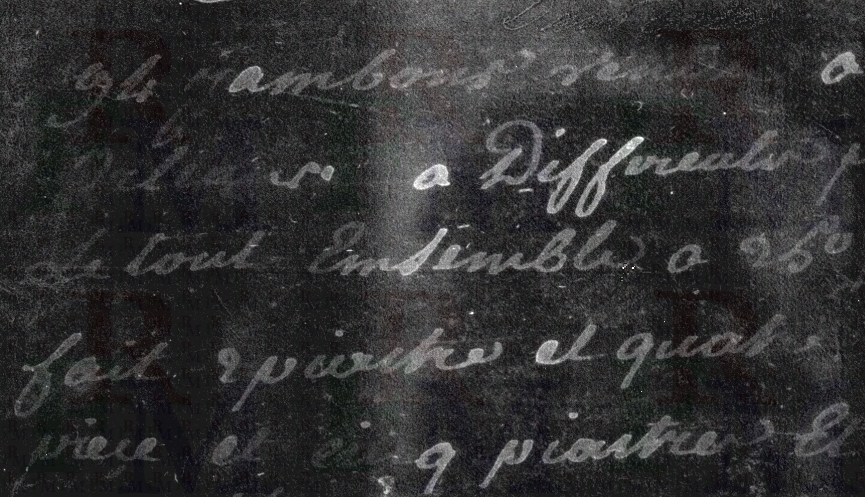} \\
(g) Ground truth of (e); Blank (Not available yet). & (h) Result of \citeS{Hedjam2014} on (e); The image is in BW10. \\\hline\hline
\end{tabular}}
\caption{The subjective evaluation of some of the CVS combinations along with the method reported in {\protect\citeS{Hedjam2014}} on the multispectral images of that paper.} \label{fig_Bin_framework_Subj2}
\end{figure*}

In Figure \ref{fig_Bin_framework_Subj2}, another subjective comparison is provided in which the performance of the proposed framework in various CVS combinations is compared with a state-of-the-art invisible text detection method \citeS{Hedjam2014}. A multispectral image from \citeS{Hedjam2014} is considered, and the outputs of the CVS ($p=0.97$, LE) configuration, the ground truth, and the result reported in \citeS{Hedjam2014} are visually compared. It is worth mentioning that the goal of the text detection method is not binarization at the pixel level, and it is more toward better visualization for human expert. This is in contrast to the 20MS dataset used in training of the proposed framework's configurations.

\subsection{The Generalization of the LE Method using \cite{Mitianoudis2014}}
\label{sec_suppl_mat_LE_plus_20}
In Section \ref{sec_Experimental_Results_Discussions}, the performance of the proposed framework has been compared with the best reported method in \cite{Mitianoudis2014} on the 9MS dataset. Various binarization kernels have been considered in the framework, such as the LE and PC binarization methods described in Section \ref{sec_Binarization_Methods}. In addition, a generalization binarization kernel, which was obtained by merging the LE method and the method of \cite{Mitianoudis2014}, has been referred to. Here, we briefly describe this combined kernel. 

Based on the observation of \cite{Mitianoudis2014}, the $7^{th}$ and $8^{th}$ bands of the MS images have been considered as the source of background information. Therefore, two variables are defined. The first variable is the background image, $I_\text{BG}$, obtained by calculating the pixel-wise mean of the $7^{th}$ and $8^{th}$ bands. The second variable is the gray image, $I$, defined in the first two steps of the binarization wrapper, i.e., applying blurring/deblurring and  the to-gray transform on the 3-band selection of the input MS image (as described in Section \ref{sec_Binarization_Wrapper_Method}). Then, $I_\text{BG}$ is adjusted to have its histogram aligned with that of $I$. Finally, $I$ is modified by removing $I_\text{BG}$ in a weighted approach to calculate the final intermediate gray image. The rest of the processes are the same as shown in Figure \ref{fig_Bin_wrapper_MS1}.

\subsection{Color to Gray: Multi-band to Single-band Conversion}
\label{sec_ToGray_Conversion}
As has been discussed in Section \ref{sec_Binarization_Wrapper_Method}, the proposed binarization wrapper requires a color-to-gray transform. Various transforms are briefly listed below:
\begin{LaTeXenumerate} \litem{Luminance} This transform attempts to encode the color information in the output gray-level image \cite{CIE20072007}:
\begin{equation*}
I_\text{lum}(x) = 1 - \Big(0.27 u_4(x)  + 0.67 u_3(x) + 0.06 u_2(x)\Big),
\end{equation*}
where $I_\text{lum}$ is the output gray image calculated in the BW10 protocol. For a traditional color image, $u_4$, $u_3$, and $u_2$ bands are equivalent to $u_\text{Red}$, $u_\text{Green}$, and $u_\text{Blue}$ bands.
\litem{Green} In this transform, only the `green' band is used: \\
$I_\text{gre}(x) = 1 - u_3(x)$.
\litem{Average} The output is the average of all three visible bands: 
$I_\text{avg}(x) = 1 - 1/3\sum_{i=2}^4 u_i(x)$.
\litem{Min-Average} A combination of the average of the bands and the band with the minimum value is used \cite{Farrahi2010}: \\
$I_\text{minavg}(x) = 1 - 1/2\left(1/3\sum_{2}^4 u_i(x) + \min_j u_j(x)\right)$.
\litem{Information Insensetive} This nonlinear conversion first rotates the color image in the RGB color space in such a way that the information difference between any of its two projections on the color axes is minimal \cite{Farrahi2013d}. Then, that projection which has the minimal intensity variation in the textual regions is selected as the output.
\end{LaTeXenumerate}

\begingroup
\bibliographystyleS{IEEEtran}
\bibliographyS{imagep}
\endgroup

\afterpage{\clearpage}
\clearpage

 \renewcommand\appendixname{ Postscript }

\setcounter{page}{1}
\setcounter{table}{0}
\setcounter{figure}{0}
\setcounter{section}{16}
\renewcommand{\thepage}{P-\arabic{page}}
\renewcommand{\thefigure}{P-\arabic{figure}}
\renewcommand{\thetable}{P-\arabic{table}}
\renewcommand{\thesection}{Postscript \Alph{section}}
\renewcommand{\thesubsection}{Postscript \Alph{section}.\Alph{subsection}}

\appendix

\section{Postscript}

\subsection{Evaluation on the MS-TEx 2015 Dataset}
\label{sec_Postscripts_Test_Dataset}
In this Postscript, the results of applying one instance of the proposed framework to the dataset of the MS-TEx 2015 contest \cite{Hedjam2015} is provided. This dataset, referred to as 10MS dataset in  \ref{sec_suppl_mat_MS_datasets_testDS}, was not available at the time of submitting this paper. We performed a comparison between CVS($p$=0.5, PC) and the contest's participants along with two state-of-the-art binarization methods: 1) The Laplacian-Energy method \cite{Howe2012} and 2) FAIR method \citeP{Lelore2013}. The full description of the contest and also its results are provided in \cite{Hedjam2015}.

\subsection{Participants in ICDAR 2015's MS-TEx 2015 Contest}
\label{sec-methods}
The full description of the contest and the participants is available in \cite{Hedjam2015}. Here, a briefed description of the participating methods is recalled for the purpose of discussing the results. Also, for clarity in referring to methods, a 4-letter code name is assigned to each method.

{\scriptsize 

\textbf{1) Computer Vision Lab. Vienna University of Technology (Markus Diem, Fabian Hollaus, and Robert Sablatnig) [DHSA]}: This method incorporates three methods for MultiSpectral Text Extraction: 1) thresholding a cleaned channel using the Lu et al. \citeP{Lu2010} binarization, 2) training an Adaptive Coherence Estimator (ACE) proposed by Scharf and Whorter \citeP{Scharf1996}, and 3) combining the cleaned channel with the mean and standard deviation images and perform a GrabCut \citeP{Rother2004}. In order to compute a cleaned channel, the background channel F8 (IR4 band) is removed from a visible channel F2 (Blue band).\footnote{The source code is available at:\\ \hreftwo{https://github.com/diemmarkus/MSTEx-CVL.git}.}

\textbf{2) Computer Vision Lab. Vienna University of Technology (Fabian Hollaus, Markus Diem, and Robert Sablatnig) [HDSA]}: In the first step of this method, the binarization method of Lu et al. \citeP{Lu2010} is applied on the Blue band (F2). The output of this method is used for the estimation of the mean spectral signature of the writing. This signature is used to train the Adaptive Coherence Estimator (ACE), which is suggested by Scharf and Whorter \citeP{Scharf1996}. The resulting binary image is then finally combined with the output of the binarization method of Lu et al \citeP{Lu2010}.\footnote{The source code is available at: \hreftwo{https://github.com/hollaus/MSTEx-CVL-matlab}.}

3)\textbf{ Document Image and Pattern Analysis (DIPA) Center, Islamabad, Pakistan (Ahsen Raza) [RAZA]}: Thís  method is based on four main steps: 1) performing image fusion using wavelet transform-based image fusion technique, 2)  performing a conditional noise removal procedure using a mix of noise removal filters, 3) performing a window- (of size $5\times 5$) based thresholding using a modified form of Niblack's thresholding technique \citeP{Niblack1985}, and 4) performing conditional noise removal followed by image cleaning based on aspect ratio of the connected components.

\textbf{4) Institute of Automation, Chinese Academy of Sciences (Alex Zhang and Cheng-Lin Liu) [ZHLI]}: The key of this method is to binarize images by a graph-based semi-supervised classification method: 1) extracting edges from the normal image (F2) using Canny edge detector, 2) coarse classification by rules, 3) fine classification by graph-based semi-supervised learning, and 4) removing the noise using F7 and F8 multispectral bands.

5) \textbf{Information Sciences Institute, University of South California (Yue Wu, Stephen Rawls, Wael Abd-Almageed, and Premkumar Natarajan) [\bf WRAN]}: this method is composed of four major stages: 1) parameter estimations, 2) feature extraction, 3)
classification, and 4) refinement. Various parameters, such as text stroke width, noise level, edge map, among others are estimated. The method uses various statistics across all the spectrum images, between pairs of the spectrum images, and also single spectrum images. Moreover, all spectrum images are binarized via a supervised base model trained on the DIBCO datasets. Finally, the method applies a learned refine classifier based on connected components analysis.

} 
\subsection{Evaluation Measures and Ranking}
In order to conform with the protocol of the MS-TEx 2015 Contest, in addition to the F-measure (FM) metrics \citeP{Rijsbergen1979,Pratikakis2012}, three other performance measures are considered in this section:\footnote{Some of these metrics are available from \citeP{Farrahi2010d}.} NRM (Negative Rate Metrics) \citeP{Ellis2002,Young2005}, DRD (Distance Reciprocal Distortion) \citeP{Wu2004,Lu2004}, and Kappa ($k$) \citeP{Cohen1968,Viera2005}. For the purpose of completeness, all measure are defined here:

\subsubsection{F-measure (FM)}
The FM metrics is a geometrical average between the precision and recall metrics:
\begin{equation}
\text{FM} =2 R  P/(R+P),
\end{equation}
where $R=\text{TP}/(\text{TP}+\text{FN})$, $P=\text{TP}/(\text{TP}+\text{FP})$ are the Recall and the Precision measures, and TP, FP, TN, and FN represent the True Positive, the False Positive, the True Negative, and the False Negative counts, respectively \citeP{Rijsbergen1979,Pratikakis2012}. For example, TP is the number of pixels that are `text' on both the binary image being evaluated $B$ and the ground truth image GT.

\subsubsection{NRM (Negative Rate Metric)}
The NRM calculates the amount of mismatch with respect to the ground truth:
\begin{equation}
\text{NRM}= \frac{1}{2}\Big(\text{R}_\text{FN}+\text{R}_\text{FP}\Big),
\end{equation}
where $\text{R}_\text{FN}=\text{FN}/(\text{TP}+\text{FN})$ and $\text{R}_\text{FP}=\text{FP}/(\text{FP}+\text{TN})$ are the False Negative Rate and the False Positive Rate, respectively \citeP{Ellis2002}.

\subsubsection{DRD (Distance Reciprocal Distortion)} 
The DRD metrics was proposed to calculate the distortion between binary images \citeP{Wu2004,Lu2004}. For all the $\cal F$ mismatching pixels, it computes the associated distortion:
\begin{equation}
\text{DRD} = \sum_{l=1}^{\cal F}\text{DRD}_l/\text{NUBN},
\end{equation}
where $\text{NUBN}$ is the number of nonuniform (not all black or white pixels) $8\times 8$ blocks in the ground truth image. $DRD_l$, which corresponds to the distortion of the $l^{th}$ mismatching pixel $x_l$ \citeP{Lu2004}, is defined to be the weighted sum of the pixels in the $5\times 5$ block of the ground truth image that differ from the value of the mismatching pixel in the binary image $B$, i.e., $B(x_l)$. $\text{DRD}_l$ can be expressed as follows:
\begin{equation}
\text{DRD}_l = \sum_{i=-2}^2\sum_{j=-2}^2 \big|GT(x_l+(i,j))-B(x_l) \big|\times W(i,j),
\end{equation}
where $W$ is a normalized weight matrix \citeP{Lu2004}.

\subsubsection{Kappa ($\kappa$)}
The Kappa ($\kappa$) coefficient \citeP{Cohen1960,Cohen1968,Viera2005}, which is well known in the domain of remotely sensed hyperspectral image classification, estimates the inter-observer reliability (reproducibility). It provides a quantitative measure of the magnitude of agreement between observers. The calculation is based on the difference between the level actual agreement (i.e., the ``observed" agreement $P_o$) compared to that level of chance-only agreement (i.e., the ``expected" agreement $P_e$):
\begin{equation}
\kappa=\frac{P_o-P_e}{1-P_e} = \frac{N_o-N_e}{N-N_e},
\end{equation}
where $N_o$, $N_e$, and $N$ are the number of matching pixels between GT and $B$, the sum of direct product of the vectors of the number of pixels in black and white classes of GT and $B$, and the total number of pixels, respectively \citeP{Cohen1960}.

{\bf Ranking:} The ranking method introduced in \citeP{Ayatollahi2013} is used. In this ranking, for every image in the dataset, best value of every metrics among all the participating methods is first determined. The participating method with this best value receives a score of 1 for the corresponding metrics, and other methods are assigned with score less than 1 depending on their performance with respect to the best value. Then, the scores of every participating method are summed together to calculate its final score $S$:
\begin{equation}
S_{k} = \sum_{i=1}^{10}\sum_{j=1}^4\Big ( \frac{\text{Best}_{i,j}}{\text{value}_{k,i,j}},\frac{\text{value}_{k,i,j}}{\text{Best}_{i,j}}\Big)_j,\quad k=1,\cdots,5,
\label{eq_S_1}
\end{equation}
where $k$ denotes the index of a particular participant, and $\text{value}_{k,i,j}$ is the value of the metrics number $j$ obtained on the test image number $i$ by the participant $k$. The operator $\big(\cdot\big)_j$ returns its first argument for those metrics $j$ that assign a lower value to a better performance (such as the DRD), and returns its second argument for those cases that the metrics $j$ shows a reverse behavior (for example, the FM). At the end, the method with the highest score $S$ is considered as the best performing method, and so on.

\begin{table}[!htb]
\small
\setlength{\tabcolsep}{2.4pt}
\begin{tabular}{||c|l||c|c|c|c||c||}\hline\hline
Rank & Method & $\text{FM}_\text{avg,1}$ & $\text{NRM}_\text{avg,1}$ & $\text{DRD}_\text{avg,1}$ & $\kappa_\text{avg,1}$ & $S$ \\\hline\hline
1$^{st}$ & 1 [DHSA] & \bf{84.87} & \bf{8.704} & \bf{3.560} & \bf{83.79} & \bf{35.12} \\\hline
2$^{nd}$ & 2 [HDSA] & 83.29 & 9.641 & 4.068 & 82.13 & 33.46 \\\hline
3$^{rd}$ & 4 [ZHLI] & 80.14 & 11.41 & 4.529 & 78.79 & 31.03 \\\hline
5$^{th}$ & 5 [WRAN] & 78.49 & 12.77 & 5.016 & 77.16 & 29.39 \\\hline
8$^{th}$ & 3 [RAZA] & 74.38 & 9.774 & 8.593 &  72.47 & 27.18 \\\hline\hline
7$^{th}$ & Howe \cite{Howe2012} & 75.96 & 8.500 & 7.806 & 74.10 & 27.50 \\\hline
6$^{th}$ & Lelore \citeP{Lelore2013} & 69.37 & {\bf 6.502} & 12.36 & 66.73 & 27.54 \\\hline\hline
4$^{th}$ & CVS($p$=0.5,PC) & 73.64 & {\bf 6.872} & 9.556 & 71.55 & {\em 29.54} \\\hline\hline
\end{tabular}
\caption{The average performance excluding the worst image of the methods against the MS-TEx 2015 dataset evaluated using four metrics. The rankings are determined using the $S$ scores.}
\label{tab_ranking}
\end{table}

\subsection{Results and Discusses on the MS-TEx 2015 Dataset}
\label{sec_res_meth_rank}
\begin{figure}[!tbh]
\centering
\scriptsize
\setlength{\tabcolsep}{1pt}
\centerline{\begin{tabular}{c}
\fbox{\includegraphics[width=2.87in, trim={0 5cm 25cm 6cm},clip]{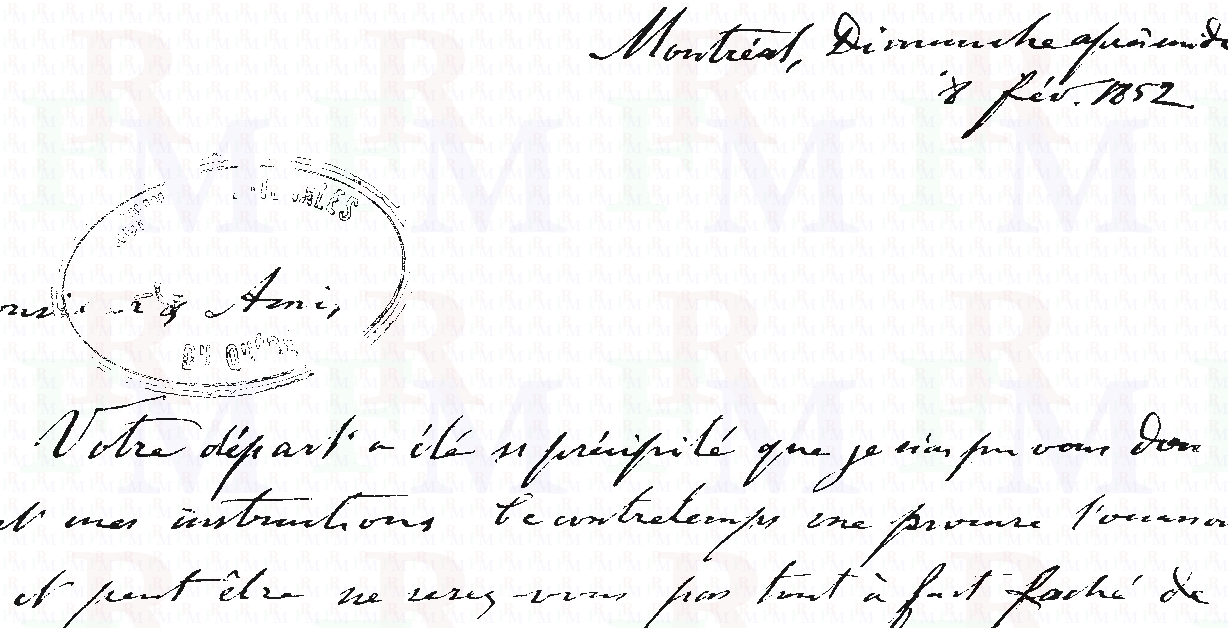}}\\
\end{tabular}}
\caption{Successful removal of the `stamp 'mark from the image {\em z31} by Method {\bf 4} [ZHLI].}
\label{fig_output_d31}
\end{figure}

\begin{figure*}[!htbp]
\centering
\centerline{\begin{tabular}{cc}
\fbox{\includegraphics[width=2.7in]{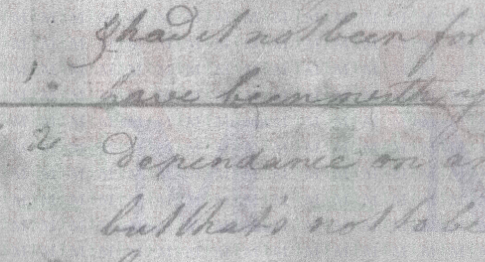}}&
\fbox{\includegraphics[width=2.7in]{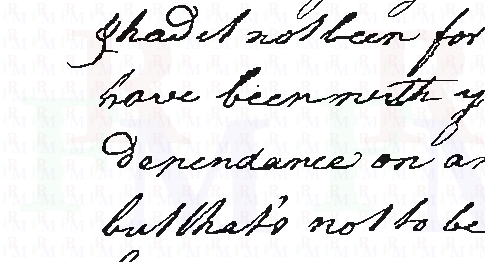}}\\
(a) & (b)\\
\fbox{\includegraphics[width=2.7in]{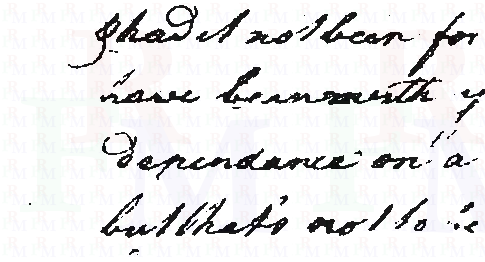}}&
\fbox{\includegraphics[width=2.7in]{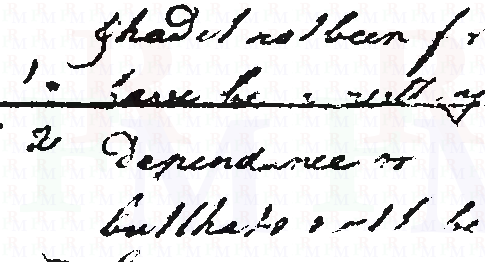}}\\
(c) & (d)
\end{tabular}}
\caption{a) The Green band (F3) of {\em z92} image of the 10MS dataset.
b) Ground truth.
c) Method {\bf 1} [DHSA].
d) The proposed framework with the PC binarization method \cite{Ziaei2014} as its kernel and $p=0.50$.}
\label{fig-output}
\end{figure*}

The performance of the methods are provided in Table \ref{tab_ranking}. Method {\bf 1} [DHSA] achieved not only the highest performance in terms of the ranking measure $S$,\footnote{The reason that the $S$ values reported in Table \ref{tab_ranking} is slightly different from those reported in \cite{Hedjam2015} is that here we considered eight binarization methods in the raking pool in contrast of seven methods that were considered in \cite{Hedjam2015}.} it also provides the best performance in terms of every individual metrics. The only exception is the NRM metrics for which FAIR and the proposed CVS methods achieve a better performance. 
 It could be argued that {\em a priori} information that the F8 band provides background noise plays a significant role in success of Method {\bf 1} [DHSA]. Similarly, Method {\bf 4} [ZHLI], which was the only method that successfully removed the `stamp' annotation (as can be seen from Figure \ref{fig_output_d31}), again uses the F7 and F8 bands in order to remove the noise. 
 Another example for the dataset is provided in Figure \ref{fig-output} (for the MS image {\em z92}).

\begin{figure*}[!tb]
\centering
\scriptsize
\setlength{\tabcolsep}{1pt}
\centerline{\begin{tabular}{cc}
\fbox{\includegraphics[width=3.1in, trim={15cm 1cm 20cm 0},clip]{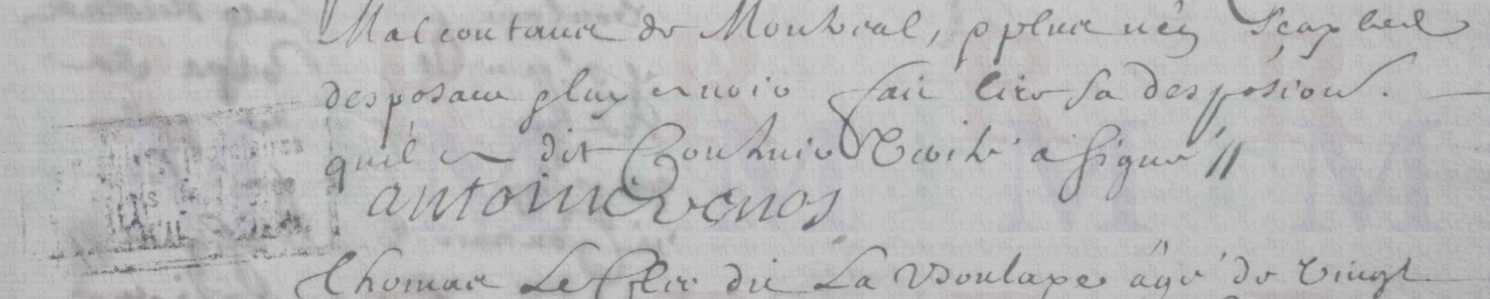}}&
\fbox{\includegraphics[width=3.1in, trim={15cm 1cm 20cm 0},clip]{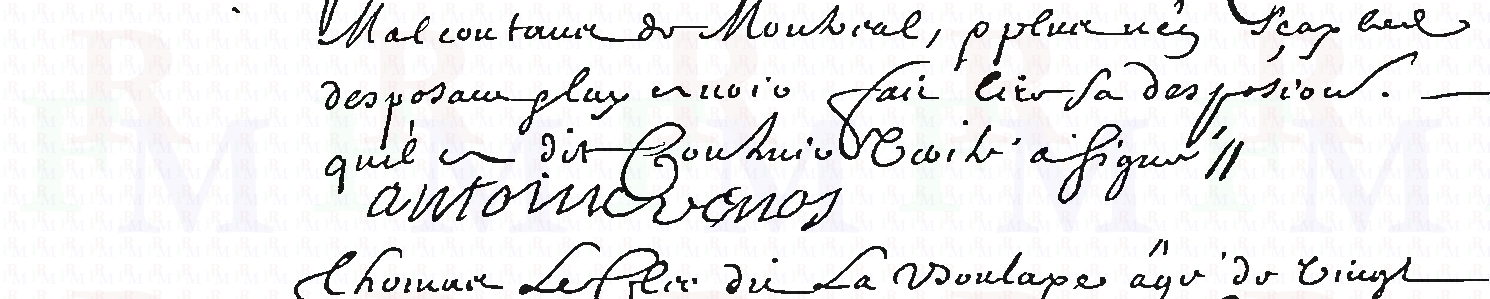}}\\
(a) Green band (F3) & (b) Ground truth \\
\fbox{\includegraphics[width=3.1in, trim={15cm 1cm 20cm 0},clip]{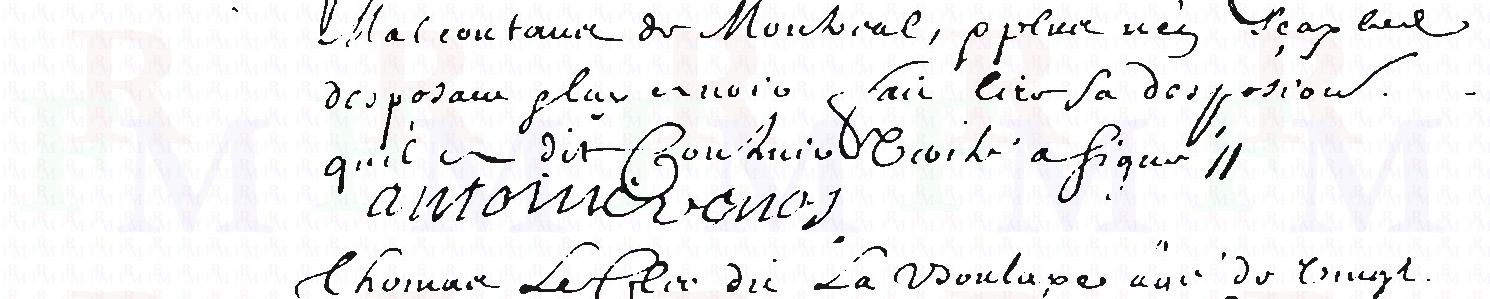}}&
\fbox{\includegraphics[width=3.1in, trim={15cm 1cm 20cm 0},clip]{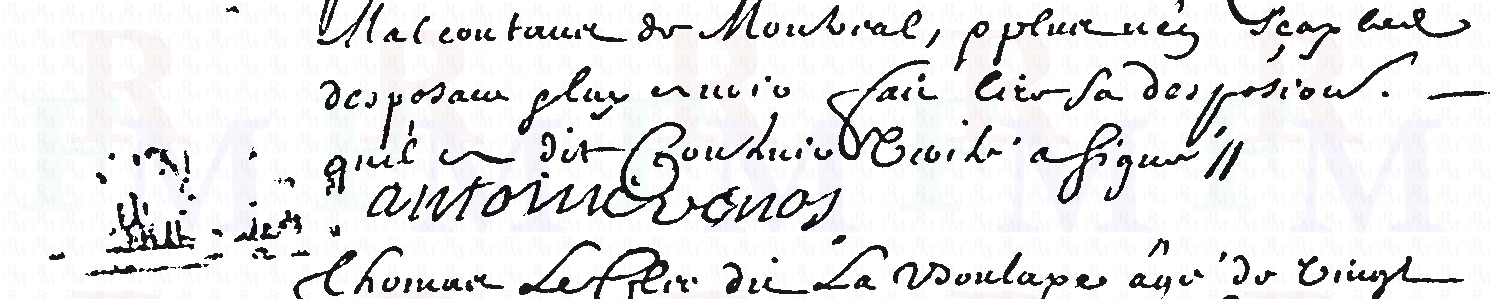}}\\
(c) Method {\bf 1} [DHSA] (NRM $=13.11$) & (d) CVS/PC/50 (NRM $=3.751$)
\end{tabular}}
\caption{The performance in terms of the NRM metrics for the image {\em z65} of the 10MS dataset.}
\label{fig_output_d65}
\end{figure*}

As can be seen from the table, the CVS method has outperformed all methods (except the FAIR method) in terms of NRM. In order to visualize this point, the image with highest difference between the NRM scores of  Method {\bf 1} [DHSA] and the CVS method is provided in Figure \ref{fig_output_d65}. The latter shows a better performance in terms of preserving connectivity of strokes. The general low performance of the proposed framework could be attributed to small size of the training dataset, especially in terms of hidden text and annotations examples, which is amplified by the blinded nature of the proposed framework that treat all the bands in the same way. In the future, we will further investigate this framework by using bigger datasets and also integrating {\em a priori} information.

\begingroup
\bibliographystyleP{IEEEtran}
\bibliographyP{imagep}
\endgroup

\end{document}